\documentclass[conference]{IEEEtran}
\IEEEoverridecommandlockouts
\usepackage{cite}
\usepackage{amsmath,amssymb,amsfonts}
\usepackage{algorithmic}
\usepackage{graphicx}
\usepackage{textcomp}
\usepackage{xcolor}
\usepackage{multirow}
\usepackage{stfloats}
\usepackage{caption}
\usepackage{hyperref}
\def\BibTeX{{\rm B\kern-.05em{\sc i\kern-.025em b}\kern-.08em
    T\kern-.1667em\lower.7ex\hbox{E}\kern-.125emX}}
\begin{document}

\title{Audio-driven Gesture Generation via Deviation Feature in the Latent Space}

% \author{\IEEEauthorblockN{1\textsuperscript{st} Jiahui Chen}
% \IEEEauthorblockA{\textit{Xiamen University} \\
% % \textit{name of organization (of Aff.)}\\
% Xiamen, China \\
% chencaffey@stu.xmu.edu.cn}
% \and
% \IEEEauthorblockN{2\textsuperscript{nd} Huan Yang}
% \IEEEauthorblockA{\textit{Giant Network} \\
% % \textit{name of organization (of Aff.)}\\
% Shanghai, China \\
% hy5ai5hy@gmail.com}
% \and
% \IEEEauthorblockN{3\textsuperscript{rd} Runhua Shi}
% \IEEEauthorblockA{\textit{Giant Network} \\
% % \textit{name of organization (of Aff.)}\\
% Shanghai, China \\
% shirunhua@ztgame.com}
% \and
% \IEEEauthorblockN{4\textsuperscript{th} Chaofan Ding}
% \IEEEauthorblockA{\textit{Giant Network} \\
% % \textit{name of organization (of Aff.)}\\
% Shanghai, China \\
% dingchaofan@ztgame.com}
% \and
% \IEEEauthorblockN{5\textsuperscript{th} Xiaoqi Mo}
% \IEEEauthorblockA{\textit{Giant Network} \\
% % \textit{name of organization (of Aff.)}\\
% Shanghai, China \\
% 134310610@qq.com}
% \and
% \IEEEauthorblockN{6\textsuperscript{th} Siyu Xiong}
% \IEEEauthorblockA{\textit{Giant Network} \\
% % \textit{name of organization (of Aff.)}\\
% Shanghai, China \\
% helenxiongsiyu@gmail.com}
% \and
% \IEEEauthorblockN{7\textsuperscript{th} Yinong He}
% \IEEEauthorblockA{\textit{Giant Network} \\
% % \textit{name of organization (of Aff.)}\\
% Shanghai, China \\
% heyinong16@gmail.com}
% }

\author{
    \IEEEauthorblockN{Jiahui Chen$^{1,2*}$, Huan Yang$^{1*}$, Runhua Shi$^{1}$, Chaofan Ding$^{1}$, Xiaoqi Mo$^{1}$, Siyu Xiong$^{1}$, Yinong He$^{1}$}
    \IEEEauthorblockA{$^1$ Ai Lab, Gaint Network, Shanghai, China}
    \IEEEauthorblockA{$^2$ Department of Digital Media Technology, Xiamen University, Xiamen, China}
    \IEEEauthorblockA{chencaffey@stu.xmu.edu.cn, hy5ai5hy@gmail.com, \{shirunhua, dingchaofan\}@ztgame.com}
}

\twocolumn[{
\renewcommand\twocolumn[1][]{#1}
\maketitle
\begin{center}
    \captionsetup{type=figure}
    \includegraphics[width=0.95\textwidth]{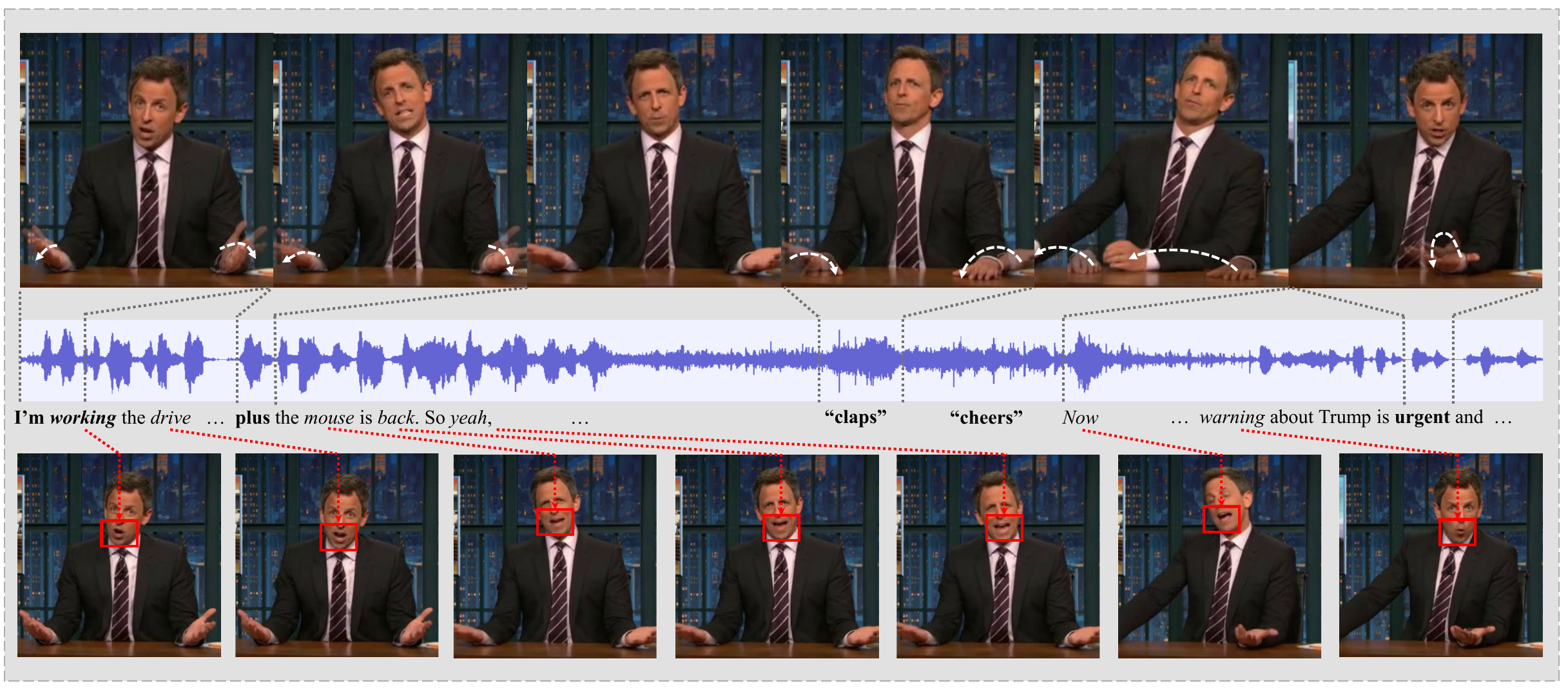}
    \captionof{figure}{Examples of our generated gesture videos. White dashed arrows indicate gestures corresponding to bold words. The red dotted boxes indicate the mouth shapes corresponding to the italicized words.}
    \label{fig:main}
\end{center}
}]

% \maketitle

\begin{abstract}
Gestures are essential for enhancing co-speech communication, offering visual emphasis and complementing verbal interactions. While prior work has concentrated on point-level motion or fully supervised data-driven methods, we focus on co-speech gestures, advocating for weakly supervised learning and pixel-level motion deviations. We introduce a weakly supervised framework that learns latent representation deviations, tailored for co-speech gesture video generation. Our approach employs a diffusion model to integrate latent motion features, enabling more precise and nuanced gesture representation. By leveraging weakly supervised deviations in latent space, we effectively generate hand gestures and mouth movements, crucial for realistic video production. Experiments show our method significantly improves video quality, surpassing current state-of-the-art techniques.
\end{abstract}

\begin{IEEEkeywords}
Co-speech gesture, audio-driven, multi-modal
\end{IEEEkeywords}

\section{Introduction}
\label{sec:intro}

Co-speech gestures, integral to human communication, enhance interaction naturalness and efficiency by complementing speech with additional information. They convey critical social cues such as personality, emotion, and subtext \cite{NGM2015Gesture}. Therefore, developing models for artificial agents that can generate contextually appropriate co-speech gestures is essential for improving human-machine interaction.

Traditional methods for generating co-speech gestures typically involve labor-intensive processes to define coherent speech-gesture pairs and associated rules \cite{Cassell_1994, Cassell_2001}. Recently, deep learning has emerged as a leading approach to automate this process. However, many existing methods focus on generating skeleton-based hand models for gesture animation \cite{Ginosar_2019, Yoon_2020}, which, while easy to generate and position, lack detailed appearance information, leading to a perceptual gap that reduces gesture realism. Recent advancements \cite{Qian_2021} have attempted to address this by applying point-level motion transformation in co-speech gesture video generation \cite{he2024co}, but these methods often introduce issues such as jitter and blurring between frames. 

To develop an effective method for automatically generating audio-driven, speech-gesture synchronized videos, we focus on motion features that capture complex gesture trajectories and detailed human appearances. Existing conditional video generation methods \cite{ruan2022mmdiffusion,Wang_2024_CVPR} typically encode videos into a latent space and use diffusion models \cite{Yan_2021,ho2022imagen,ceylan2023pix2video} for content generation. However, these general-purpose approaches often yield motion videos lacking in realism and fine-grained detail, and they demand significant computational resources. Moreover, prior studies on co-speech gestures \cite{Chaitanya_2020,Yoon_2020,Ginosar_2019} have mainly concentrated on hand and arm gestures using fully supervised learning \cite{corona2024vlogger}, which relies heavily on large-scale, labor-intensive annotations.

To address these challenges, we propose a novel approach for generating co-speech gesture videos. First, a deviation module is proposed to generate latent representation of both foreground and background in the motion for co-speech video generation. This deviation module is consisted of latent deviation extractor, a warping calculator and a latent deviation decoder. Second, a corresponding weakly supervised learning strategy is proposed to achieve the latent representation of deviation for high-quality video generation. In the evaluation experiments, the results demonstrate high-quality co-speech video generation and outperform the state-of-the-art models in the evaluation metrics.

%First, we design a latent motion extractor to capture both gesture motion and appearance information from video inputs. Specifically, we map the driving images into a latent space to extract motion features, which are then used to train the generative model. Next, we employ a diffusion model that leverages self-attention and cross-attention mechanisms to more effectively capture the temporal correlations between speech and motion. Additionally, we condition the diffusion model on preceding frames of the current motion feature, effectively addressing issues of jitter and blurring between frames. Finally, we introduce a latent deviation decoder that reconstructs the final image, following the deformation of the decoded motion features.

In summary, our main contributions are as follows: 1) We propose a deviation module in order to produce pixel-level deviation of both fore-ground and back-ground in the co-speech video generation. This deviation module is consisted of three parts including a latent deviation extractor, a warping calculator and a latent deviation decoder. 2) The latent representation of deviation is learned in a proposed weakly supervised learning strategy. For both training and inference, it enable the representation of pixel-level deviation to be applied in the end to end video generation.

\begin{figure*}[ht!]
    \centering
    \includegraphics[width=0.98\linewidth]{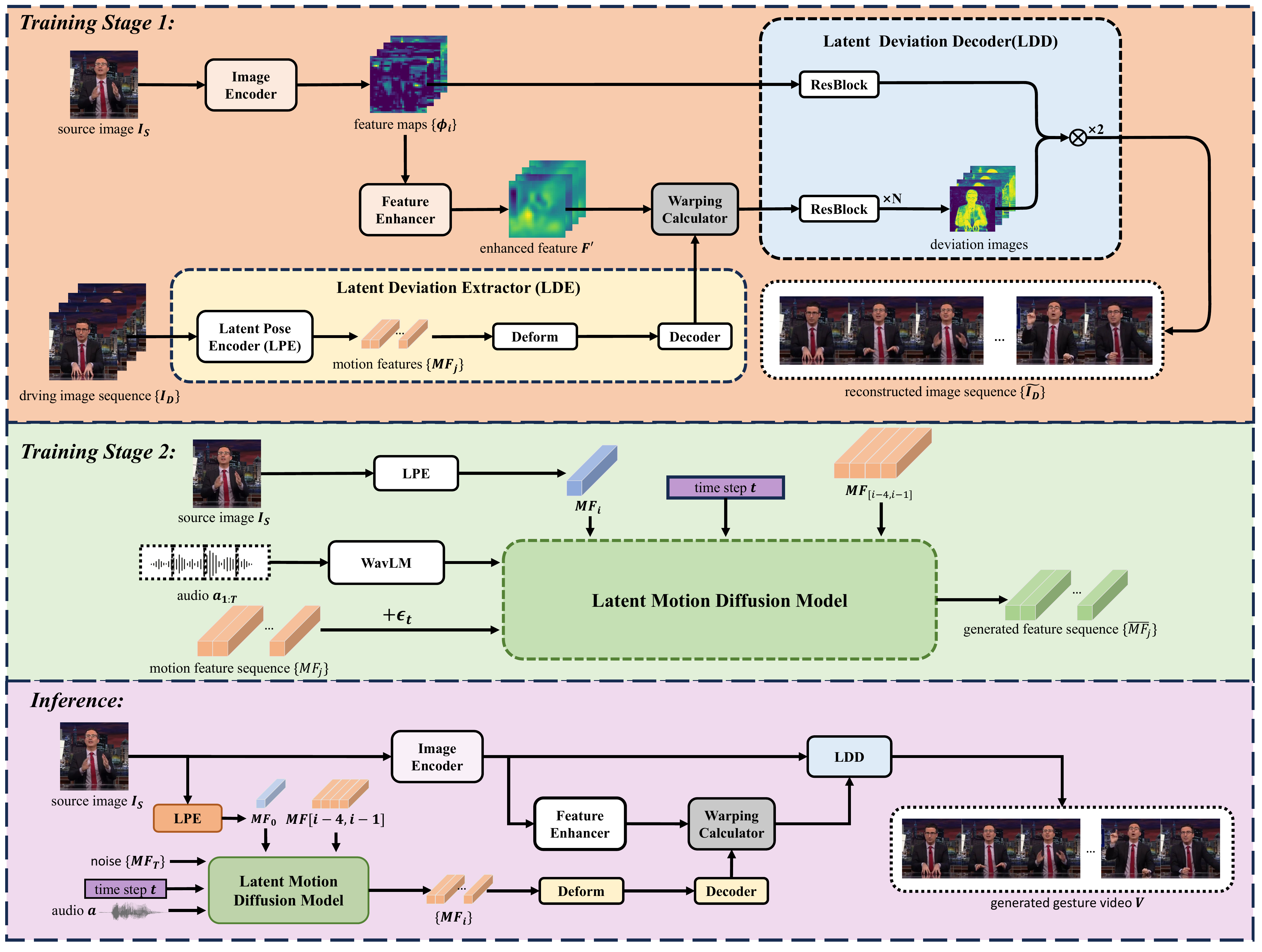}
    \caption{Co-speech gesture video generation pipeline of our proposed method consists of three main components: 1) the latent deviation extractor (yellow) extracts motion features from videos and predicts optical flow; 2) the latent deviation decoder (blue) applies deviation to the motion optical flow and decodes the image features to reconstruct the image; 3) the latent motion diffusion (green) generates motion features based on the given speech.}
    \label{fig:overview}
\end{figure*}

\section{Methodology}

We propose a novel method for generating co-speech gesture videos, utilizing a full scene deviation, produces co-speech gesture video $V$ (i.e. image sequence) that exhibit natural poses and synchronized movements. The generation process takes as input the speaker's speech audio $a$ and a source image $I_S$. An overview of our model is shown in Fig.~\ref{fig:overview}. 

We structured the training process into two stages. In the first stage, a driving image $I_D$ and a source image $I_S$ are used to train the base model. In one aspect, the proposed latent deviation module consisting of latent deviation extractor, warping calculator and latent deviation decoder is trained under weakly supervision. In another aspect, other modules in the base model is trained under full supervision. In the second stage, the motion features, consisting of $MF_i$, $\widetilde{MF}_[i-4,i-1]$, and the noise-added motion feature sequence $\{MF_j\}$, are used to train the latent motion diffusion model. In the following parts, we will introduce the two stages of training part and the inference part in detail. 

\subsection{Stage 1: Base Model Learning}

% \subsubsection{Image encode} First, we input the source image $I_S \in \mathbb{R}^{H\times W \times 3}$ into the image encoder $\mathcal{E}$ to obtain the feature $F$ and feature maps $\{\phi_{i\in N}\}$, as the layer-by-layer feature maps provide additional background details for subsequent steps. Directly inputting shallow spatial features $F$ into the warping calculator increases the training difficulty and makes it more likely to decode blurred images with artifacts after applying optical flow transformations. Therefore, we use a feature enhancer to map the shallow spatial features to a higher-level space, achieving better results. The equation is as follows:
% \begin{equation}
%     F' = \frac{F - \bar{F}}{\sqrt{\sigma^2+\epsilon}}\gamma + \beta,
% \end{equation} where $\bar{F}$ and $\sigma$ represent the mean and standard deviation of the features, while $\gamma$ is the scaling factor, and $\beta$ is the bias.

\subsubsection{Weakly Supervised Deviation} To extract human motion information from image sequences, we propose a latent space motion feature extractor that differs from the MRAA used in ANGIE \cite{liu2022audio} and the TPS transformation-based motion decoupling module in S2G-MDDiffusion \cite{he2024co}. Our approach involves extracting motion features $MF$ from the driving image $I_D \in \mathbb{R}^{H\times W \times 3}$ , followed by applying a nonlinear transformation and a motion decoder $\psi$ to generate an optical flow map. This map is then combined with the enhanced source image features $F'$ using a warping calculator $\mathcal{W}$, producing a motion-informed feature map $F'_{\mathcal{W}}$. Finally, our latent deviation decoder (LDD) processes $F'_{\mathcal{W}}$ alongside the source image's feature map $F$ through a scene deviation mechanism to reconstruct the gesture image $\widetilde I_D$.

% LPE 描述
\paragraph{Latent Deviation Extractor} To generate natural motion-transformed images, we design a Latent Pose Encoder (LPE) that extracts a latent motion feature $MF \in \mathbb{R}^{1 \times K}$. This feature undergoes a nonlinear pose transformation before decoding. The compact latent representation of pose transformations enables the generation of concise optical flow, effectively driving the image transformation.

\paragraph{Warping Calculator} To effectively integrate motion information into the source image $I_S$, we first encode it to enhance its feature maps. We then apply a warping function $\mathcal{W}(\cdot)$ using optical flow rotation $\mathbf{R}$ and translation $\mathbf{T}$ matrices of optical flow to obtain the deformed features $F'_\mathcal{W}$. This enhancement process amplifies key features while reducing background noise, resulting in a clearer representation of critical information.

\paragraph{Latent Deviation Decoder} Due to the occlusions and misalignments between $I_S$ and $I_D$, directly decoding after the warping operation often fails to achieve effective image reconstruction. Inspired by~\cite{zhao2022thin,corona2024vlogger}, we add an full scene deviation $\delta_F$ into the decoder during the decoding process, which improves the accuracy of the reconstructed image $\widetilde I_D$. The full scene deviation formula is as follows:
\begin{equation}
    \delta_F = L \frac{1}{1 + e^{-(wF'_\mathcal{W} + b)}},
\end{equation} which is a variant of the sigmoid function. 
\begin{figure}
    \centering
    \includegraphics[width=\linewidth]{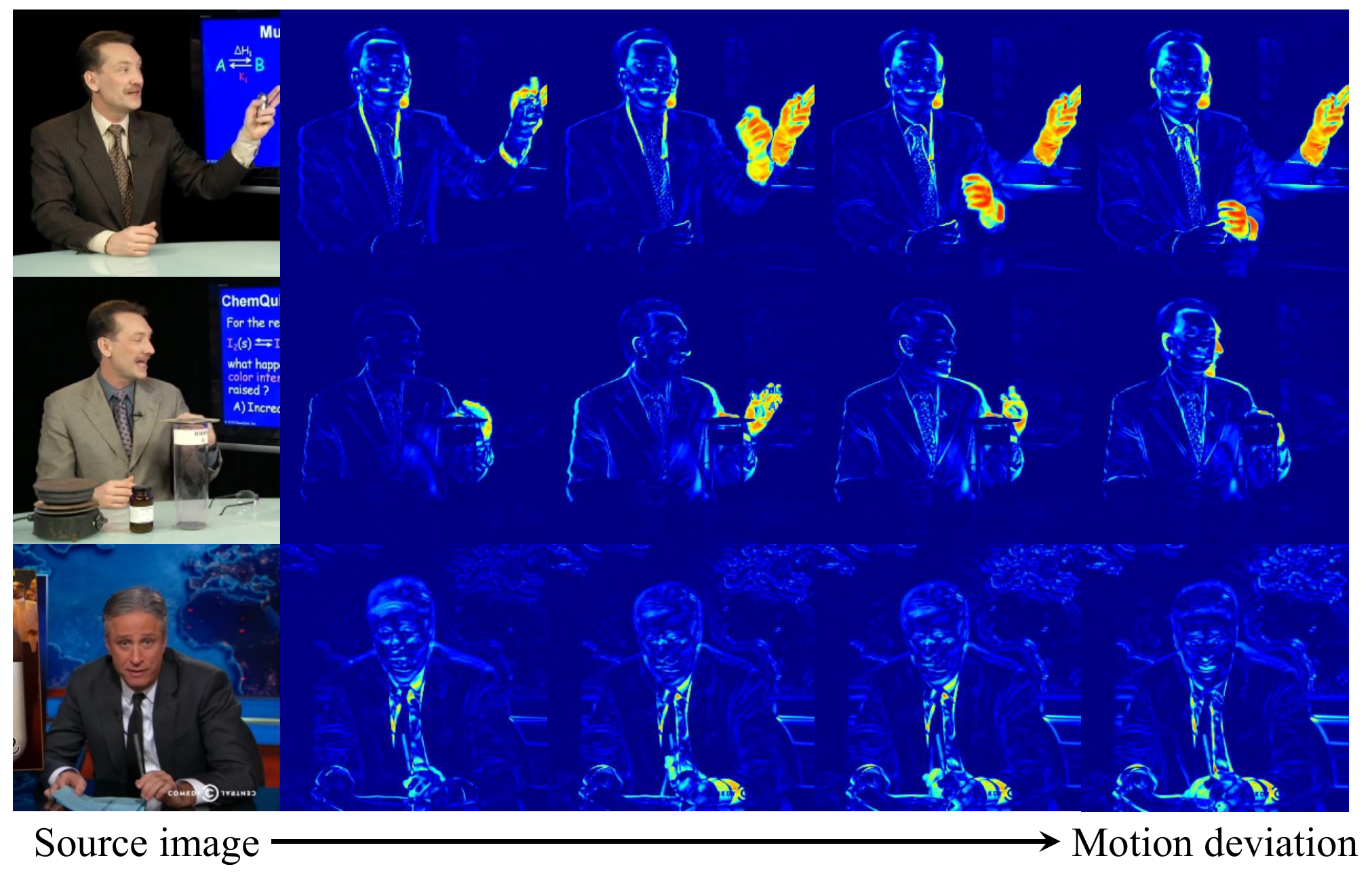}
    \caption{The deviation in latent representation. Heatmap shows the deviation of gestures and other movements in the latent space.}
    \label{fig:deviation} 
\end{figure}
Next, we interpolate and decode the deviation $\delta_F$ together with the feature $F$. This approach allows the motion features to be smoothly interpolated onto the source image features, resulting in more natural outcomes. The decoding process is as follows:
\begin{equation}
    \mathbf{z} = \delta_F F + (1 - \delta_F) U(F), 
\end{equation} where $U(\cdot)$ denotes image decoder. Moreover, we design a nonlinear activation function that can adjust the impact of negative input values. By tuning the parameter $c_{\lambda}$, we can control the contribution of negative input values to the output.
% % 删除激活函数
% \begin{equation}
%     \mathbf{a} = max(0, \mathbf{z}) + c_{\lambda} min(0, \mathbf{z}).
% \end{equation}

\subsubsection{Training} 
In the training process of co-speech gesture video generation, unlike vlogger~\cite{corona2024vlogger} which applies a large-scale of annotation data for supervised dense representation of motion, we directly apply image reconstruction loss without dense/sparse supervised annotation of motion~\cite{bhat2023self}. Following~\cite{Siarohin2021motion,bhat2023self}, we utilize a pre-trained VGG-19~\cite{johnson2016perceptual} network to calculate the reconstruction global loss $\mathcal{L}_{per\_glo}$ between the reconstructed image $I_D$ and the generated image $\widetilde I_D$ across multiple resolutions.
% \begin{equation}
%     \mathcal{L}_{per\_glo} = \sum_{j} \sum_{i} c_i |V_{i}(I_{Dj}) -V_{i}(\widetilde I_{Dj})|,
% \end{equation}
% where $V_i$ is the $i$-th layer of the pre-trained VGG-19 network, and $j$ represents that the image is downsampled $j$ times.

Additionally, we calculated the local loss $\mathcal{L}_{per\_loc}$ of the reconstructed image, which consists of hand loss $\mathcal{L}_{hand}$ and face loss $\mathcal{L}_{face}$, based on the VGG-16 model. To ensure realism, we employed a patch-based discriminator~\cite{quan2022image} and trained it using the GAN discriminator loss $\mathcal{L}_{discr}$ for adversarial training. Both the ground truth images and the generated fine images are converted into feature maps, where each element is classified as real or fake. Therefore, the final loss is the following:
\begin{equation}
    \mathcal{L}_1 = \lambda_{per}\mathcal{L}_{per} + \lambda_{GAN}\mathcal{L}_{GAN} + \lambda_{discr}\mathcal{L}_{discr}.
\end{equation}

\subsection{Stage 2: Latent Motion Diffusion}

Given that the extracted motion features $MF$ reside in the latent space, we adopted and modified the latent motion diffusion model proposed by \cite{he2024co} for our generation task. Our diffusion model accepts five inputs: time step $t$, audio $a$, the noisy motion feature sequence $\{MF_{j\in M}\}$, the motion feature ${MF}_i$ of the source image, and the predicted motion features $\widetilde {MF}_{[i-4, i-1]}$ from the previous four frames. The model then predicts a clean motion feature sequence $\{\widetilde {MF}_{i\in M}\}$ from noised $\{MF_{j\in M}\} + \epsilon_t$ and condition $c = (a, \widetilde {MF}_{[i-4, i-1]}, {MF}_i)$.

\subsubsection{Feature Priors} To address issues like frame skipping, blurring, and jitter in current gesture video generation, we introduce motion feature priors to enhance video stability. These priors include weakly-supervised deviation features and fully-supervised features. Specifically, the weakly-supervised deviation features are the predicted motion features $\widetilde {MF}_{[i-4, i-1]}$, while the fully-supervised features are the motion features $\{MF_{j\in M}\}$ encoded by LPE. By incorporating both the current frame's motion features and the predicted features from the preceding four frames as conditional inputs, we leverage the inherent temporal continuity between frames. This approach improves the coherence and realism of the predicted motion sequence, effectively mitigating issues such as frame skipping. For the $0-th$ frame, where preceding motion features are unavailable, we initialize the features of the preceding four frames to zero.

\subsubsection{Loss} We divide the loss into three components. The first is a motion feature loss calculated with MSE, which constrains the movements of the hands, lips, and head, promoting naturalness and coherence. Additionally, the loss includes implicit velocity and implicit acceleration losses~\cite{li2022cvpr} to prevent the overall motion from being too rapid. The final training loss is as follows:
\begin{equation}
    \mathcal{L}_{diff} = \mathcal{L}_{MF} + \lambda_{im\_vel}\mathcal{L}_{imp\_vel} + \lambda_{imp\_acc}\mathcal{L}_{imp\_acc}.
\end{equation}

\subsection{Inference}

As shown in Fig.~\ref{fig:overview}, given a source image $I_S$ and audio $a$ as inputs, motion feature ${MF}_0$ will be 
first encoded with LPE. Conditioned on ${MF_0}$ and extracted audio features $a$, we randomly sample a Gaussian noise ${MF}_T \in \mathbb{R}^{M \times 1 \times K}$ from $\mathcal{N}(0, \textbf{I})$ and denoise it. After $T$ steps, we obtain a clean sample x0. Repeating this procedure, we can get a consistent and coherent long sequence of motion features $\{\widetilde {MF}_{i\in M}\}$. For each frame $\widetilde {MF}_i$, we use a nonlinear transformation and a pose decoder to obtain optical flow. This is then fed into a warping calculator that incorporates enhanced feature $F'$ of the source image to produce the motion feature map of the source image. Finally, we apply deviation and decoding to the motion feature map and source image's feature map, converting them into a fine-grained result. All frames are then assembled to form a complete co-speech gesture video.

\begin{table*}[ht!]
    \centering
    \begin{tabular}{c c c c c c c c c}
    \multirow{2}{*}{Name} & \multicolumn{4}{c}{Objective evaluation} & \multicolumn{4}{c}{Subjective evaluation} \\
    \cline{2-9}
    \multirow{2}{*}{} & FGD $\downarrow$ & Div. $\uparrow$ & BAS $\uparrow$ & FVD $\downarrow$ & Realness $\uparrow$ & Diversity $\uparrow$ & Synchrony $\uparrow$ & Overall quality $\uparrow$ \\
    \hline
    % Ground Truth (GT) & 8.976 & 5.911 & 0.1498 & 1852.86 & 4.70$\pm$ 0.07 & 4.45$\pm$0.05 & 4.66$\pm$0.08 & 4.70$\pm$0.07 \\
    % \hline
    ANGIE & 55.655 & 5.089 & \textbf{0.1513} & 2846.13 & 2.01$\pm$0.07 & 2.38$\pm$0.07 & 2.11$\pm$0.07 & 1.98$\pm$0.08 \\
    MM-Diffusion & 41.626 & 5.189 & 0.1021 & 2802.18 & 1.63$\pm$0.07 & 1.93$\pm$0.08 & 1.57$\pm$0.09 & 1.44$\pm$0.08 \\
    S2G-MDDiffusion & \underline{18.745} & \underline{5.702} & {0.1254} & \underline{2019.585} & \underline{2.78$\pm$0.07} & \underline{3.21$\pm$0.08} & \underline{3.32$\pm$0.08} & \underline{2.89$\pm$0.07} \\
    TANGO & 32.494 & 4.317 & 0.0927 & 3106.470 & {2.15$\pm$0.07} & {1.41$\pm$0.08} & {2.12$\pm$0.06} & {1.75$\pm$0.06} \\
    Ours & \textbf{15.896} & \textbf{6.068} & \underline{0.1281} & \textbf{1861.101} & \textbf{3.82$\pm$0.07} & \textbf{3.94$\pm$0.07} & \textbf{4.11$\pm$0.06} & \textbf{3.93$\pm$0.05} \\
    \hline
    \end{tabular}
    \caption{Quantitative results on test set. Bold indicates the best and underline indicates the second. For ANGIE~\cite{liu2022audio} and MM-Diffusion~\cite{ruan2022mmdiffusion}, we cited the results in the S2G-MDDiffusion~\cite{he2024co} paper. For S2G-MDDiffusion, we used the official open source code.}
    \label{tab:quantitative}
\end{table*}

\begin{table*}[ht!]
    \centering
    \resizebox{0.95\textwidth}{!}{
    \begin{tabular}{c c c c c c c c c c c c}
        \multirow{2}{*}{Name} & \multirow{2}{*}{Image quality $\uparrow$} & \multirow{2}{*}{Motion smoothness $\uparrow$} & \multicolumn{3}{c}{Hand gesture} & \multicolumn{3}{c}{Lip movement} &  \multicolumn{3}{c}{Full image} \\
        \cline{4-6} \cline{7-9} \cline{10-12}
        & & & PSNR $\uparrow$ & SSIM $\uparrow$ & LPIPS $\downarrow$ & PSNR $\uparrow$ & SSIM $\uparrow$ & LPIPS $\downarrow$ & PSNR $\uparrow$ & SSIM $\uparrow$ & LPIPS $\downarrow$ \\
        \hline
        S2G-MDDiffusion & 0.624 & 0.995 & 22.37 & 0.625 & 0.106 & 23.84 & 0.689 & 0.092 & 29.39 & 0.952 & 0.034 \\
        Ours & \textbf{0.674} & \textbf{0.997} & \textbf{23.91} & \textbf{0.756} & \textbf{0.054} & \textbf{29.10} & \textbf{0.882} & \textbf{0.038} & \textbf{31.79} & \textbf{0.976} & \textbf{0.018} \\
        \hline
    \end{tabular}
    }
    \caption{Comparison of gestures and mouth movements based on common image quality metrics.}
    \label{tab:comparisons2}
\end{table*}

\section{Experiments}

\begin{figure*}[ht!]
    \centering
    \includegraphics[width=\linewidth]{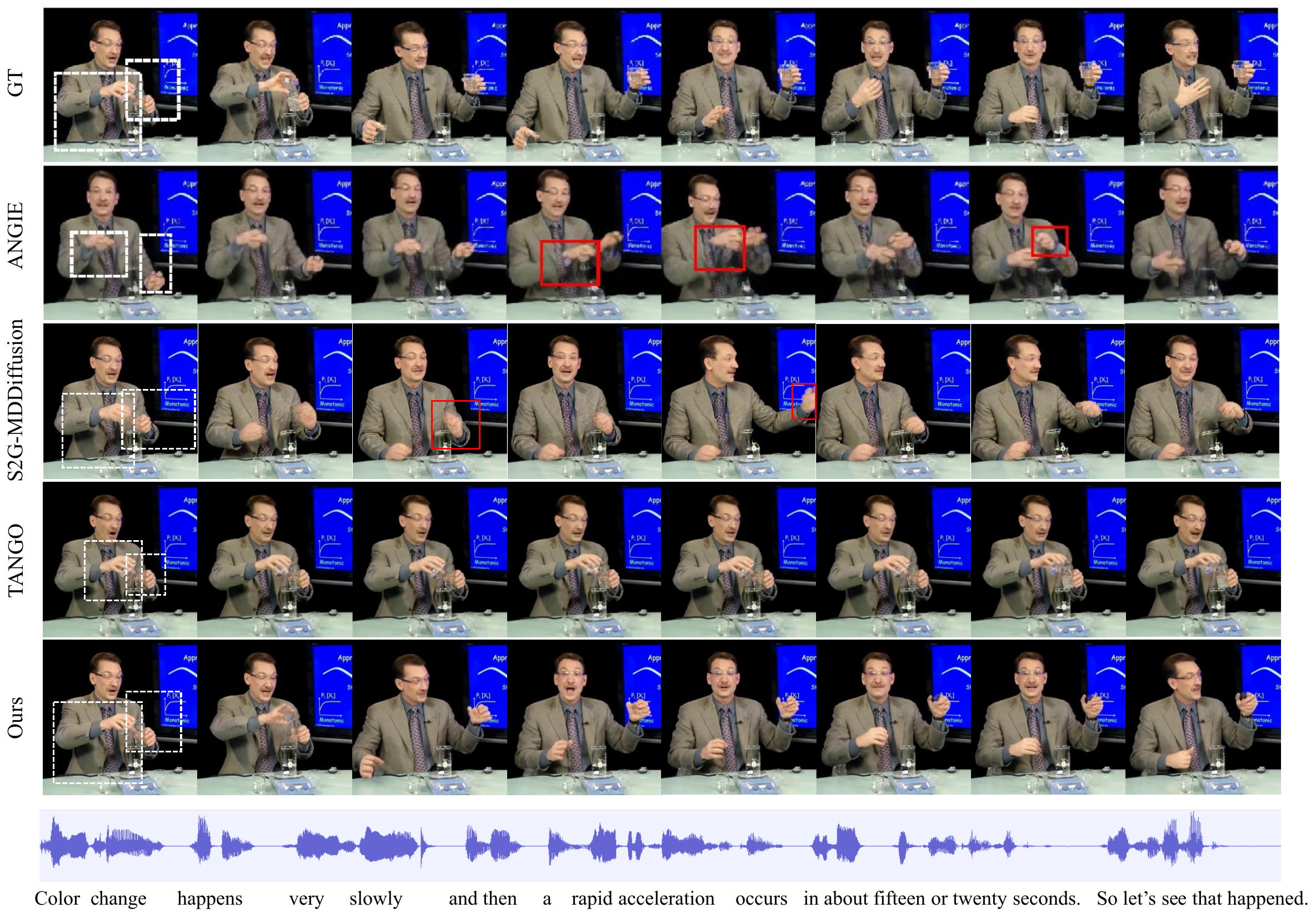}
    \caption{Visual comparison with SOTAs. Our method generates gestures with more extensive accurate motions (dashed boxes), matching audio and semantics. Red boxes indicate unrealistic gestures generated by ANGIE~\cite{liu2022audio}, S2G-MDDiffusion~\cite{he2024co} and TANGO~\cite{liu2024tango}.}
    \label{fig:comparisons1}
\end{figure*}

\subsection{Implementation Details}

\textbf{Dataset.} Our experimental data comes from the PATS dataset~\cite{Chaitanya_2020}, which consists of gesture clips with aligned audio and text transcriptions, including approximately 84,000 clips from 25 speakers, with an average length of 10.7 seconds, totaling about 250 hours. We preprocessed the video data for four individuals: Chemistry, Oliver, Jon, and Seth and following the standard process from S2G-MDDiffusion~\cite{he2024co}.  
%The videos were converted to 25 fps, and the timing information for the characters was sourced from PATS. The box information for the characters was derived from S2G-MDDiffusion~\cite{he2024co}, specifically calculating the frame width and left edge ratio. The cropped videos were then scaled to 256×256, with the audio extracted from the videos and saved in wav format at a sampling rate of 44100 Hz, with a maximum video duration of 15 seconds.

\textbf{Evaluation metrics.} We employed 1) \textbf{Fréchet Gesture Distance (FGD)} to evaluate the distance between the distributions of real and synthetic gestures. An auto encoder was trained on the PATS image dataset, and the encoder was used to calculate the Fréchet distance between real and synthetic gestures in the feature space. We used 2) \textbf{Fréchet Video Distance (FVD)} and 3) \textbf{Diversity (Div.)} to assess the overall quality of gesture videos and the diversity of the generated gestures. Diversity measures the differences in gestures corresponding to different audio inputs in the latent space. Also,  we compute the average distance between closest speech beats and gesture beats as 3) \textbf{Beat Alignment Score (BAS)} following~\cite{Li_Yang_Ross_Kanazawa_2021}. Besides, we also apply PSNR, SSIM and LPIPS~\cite{corona2024vlogger} for the evaluation of generated hand gesture, lip movement and full scene, and we calculate image quality following~\cite{ke20221musiq} and motion smoothness following~\cite{huang2023vbench}. 

% The pre-processing method for preparing the input depth image includes first cropping the hand area from a depth image similar to\cite{DBLP:journals/corr/abs-1708-08325}, and then resizing it to a fixed size of 128 $\times$128.  
% \subsection{ Datasets and Evaluation Metrics}
%  \textbf{ICVL Dataset.} The ICVL dataset \cite{Tang2014Latent} provides $22K$ and $1.6K$ depth frames for training and testing, respectively.
 
%  \textbf{NYU Dataset.} The NYU dataset \cite{2014Real} is captured from three different views with Microsoft Kinect sensor. 
 
%  \textbf{Evaluation metrics.} We use the commonly used metrics for the evaluation of 3D hand pose estimation: the mean distance error (in mm). 

\subsection{Comparison with Existing Methods}

We compared our method with gesture video generation methods ANGIE~\cite{liu2022audio}, MM-Diffusion~\cite{ruan2022mmdiffusion}, S2G-MDDiffusion~\cite{he2024co}, and TANGO~\cite{liu2024tango}, a method that builds on Gesture Video Reenactment, which represents video frames as nodes and valid transitions as edges.

\begin{figure}[ht!]
    \centering
    \includegraphics[width=\linewidth]{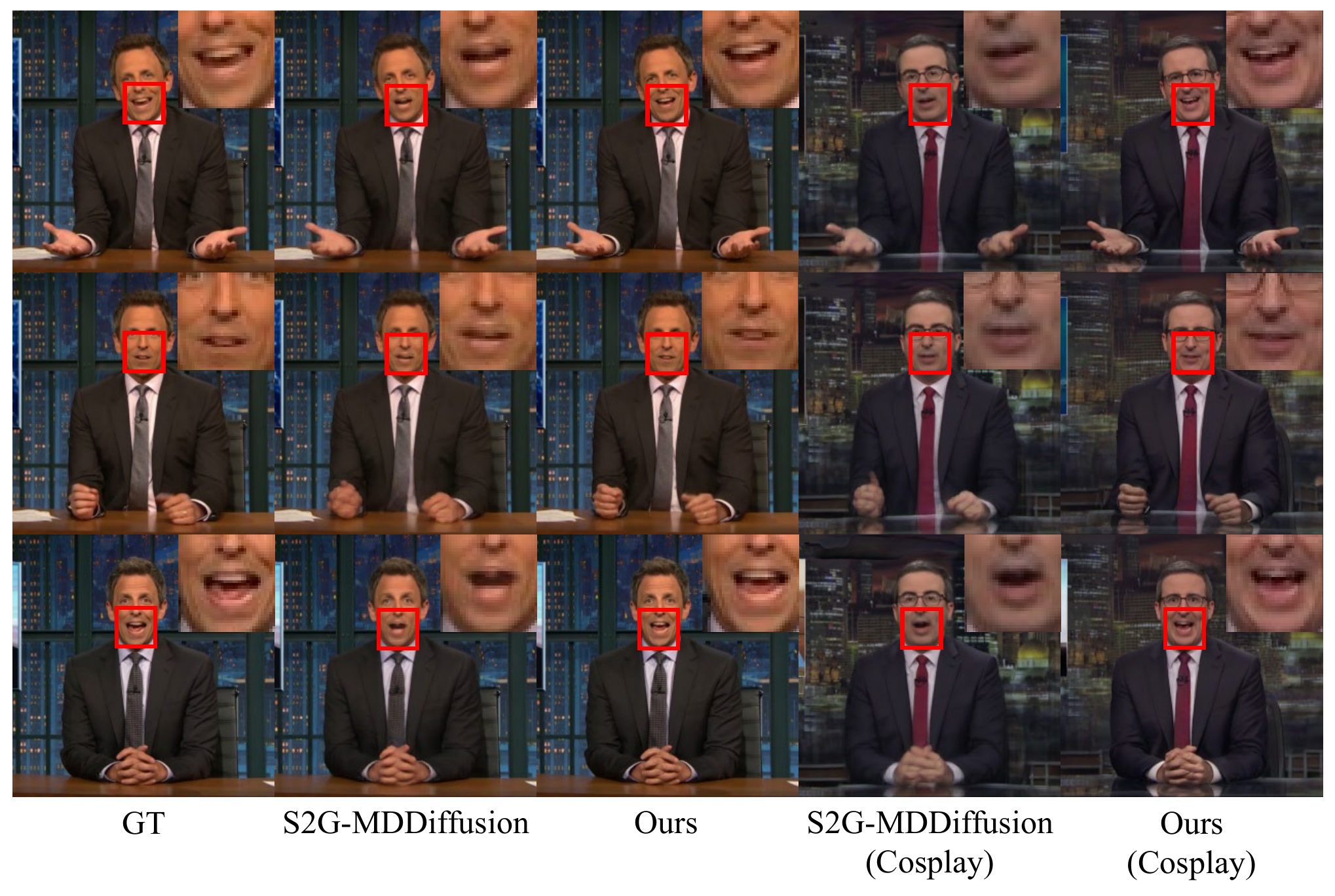}
    \caption{Comparison of drivers of the same and different people. The gesture videos we generate are not only better in gesture expression, but also more natural in facial expressions and lip movements.}
    \label{fig:comparisons3}
\end{figure}

In Fig.~\ref{fig:comparisons1}, we compare the gesture range and hand clarity across identical frames from original video. Our method mitigate abnormal deformations in comparison with the SOTA models. We observed that the TANGO method relies on video keyframes for interpolation, which limits its ability to generate novel content. Consequently, it can only produce videos of repetitive actions lasting approximately one second. In Fig.~\ref{fig:comparisons3}, we also compare generation of hand gesture, facial expression and lip movements, our method also improves the quality of these parts. In detail, the proposed method mitigates jitter, blurred region(hands/faces/lips) and other abnormal deformations. The quantitative results are shown in Table~\ref{tab:quantitative} and~\ref{tab:comparisons2}. Our method outperforms existing approaches in terms of image quality, motion smoothness, FGD, Diversity, and FVD metrics, PSNR, SSIM and LPIPS. 

% \begin{figure}[ht!]
%     \centering
%     \includegraphics[width=\linewidth]{images/comparisons2.pdf}
%     \caption{Visualization results of fine-grained hand variations. The gesture videos we generate are clearer, more reasonable, more diverse and more natural in the same frame.}
%     \label{fig:comparisons2}
% \end{figure}

\subsection{User Study}
To better understand the subjective performance of our method, we conducted a user study inspired by S2G-MDDiffusion~\cite{he2024co} to evaluate the similarity between gesture videos generated by each method and the ground truth. We selected 20 generated videos from the PATS test set, ranging from 4.5 to 12.4 seconds in length. Twenty participants were invited to provide Mean Opinion Scores (MOS) across four dimensions, including 1) \textbf{Realness}, 2) \textbf{Diversity} of gesture, 3) \textbf{Synchrony} between speech and gestures, and 4) \textbf{Overall quality}, with scores ranging from 1 to 5 in 0.5 increments, where 1 indicates the worst and 5 indicates the best. The results show that our method achieved satisfactory user ratings across all four dimensions.

\begin{table*}[ht!]
    \centering
    % \resizebox{\textwidth}{15mm}{
    \begin{tabular}{c c c c c c c c c}
    \multirow{2}{*}{Name} & \multicolumn{4}{c}{Objective evaluation} & \multicolumn{4}{c}{Subjective evaluation} \\
    \cline{2-9}
    \multirow{2}{*}{} & FGD $\downarrow$ & Div. $\uparrow$ & BAS $\uparrow$ & FVD $\downarrow$ & Realness $\uparrow$ & Diversity $\uparrow$ & Synchrony $\uparrow$ & Overall quality $\uparrow$ \\
    \hline
    w/o Dev. in LDD & {17.356} & 5.173 & 0.1212 & {1910.462} & 3.03$\pm$0.08 & 3.62$\pm$0.08 & 3.62$\pm$0.07 & 3.29$\pm$0.08 \\
    
    w/o $F'$ & {17.567} & {5.405} & \textbf{0.1285} & 2017.193 & 3.62$\pm$0.06 & 3.81$\pm$0.09 & 3.80$\pm$0.08 & 3.67$\pm$0.08 \\
    
    w/o Mo Dec. & \underline{16.474} & \underline{5.450} & 0.1207 & \underline{1898.149} & \underline{3.66$\pm$0.07} & \underline{3.85$\pm$ 0.09} & \underline{3.86$\pm$0.06} & \underline{3.77$\pm$0.06} \\
    \hline
    Ours & \textbf{15.896} & \textbf{6.068} & \underline{0.1281} & \textbf{1861.101}  & \textbf{3.87$\pm$0.07} & \textbf{3.86$\pm$0.08} & \textbf{4.08$\pm$0.07} & \textbf{3.86$\pm$0.08} \\
    \hline
    \end{tabular}
    % }
    \caption{Ablation study results. Bold indicates the best and underline indicates the second. 'w/o' is short for 'without'.}
    \label{tab:ablation}
\end{table*}

% \begin{table}[ht!]
%     \centering
%     \begin{tabular}{c c c c}
%         Name & PSNR & SSIM & LPIPS \\
%         \hline
%         S2G-MDDiffusion & 23.84 & 0.689 & 0.092 \\
%         Ours & 29.10 & 0.882 & 0.038 \\
%         \hline
%     \end{tabular}
%     \caption{Caption}
%     \label{tab:comparisons3}
% \end{table}

\subsection{Ablation Study}
We conducted an ablation study to demonstrate the effectiveness of different components in our method. The results are shown in Table~\ref{tab:ablation}. We examined the effectiveness of the following components: 1) the deviation in the LDD, 2) the enhanced feature, and 3) the motion decoder.

Supported by the result in Table~\ref{tab:ablation}, removing the deviation from the Latent Deviation Decoder (LDD) significantly reduces diversity, as this deviation is crucial for calibrating the range of motion features. The absence of enhanced features leads to decreased performance in FGD and FVD, since these features encapsulate higher-level semantic information essential for generating high-quality gesture videos. Conversely, the lack of enhanced features increases BAS by allowing stronger expression of audio features during decoding. Additionally, the motion decoder enhances FGD, diversity, and FVD.

\section{Conclusion}
In this paper, we present a weakly-supervised approach for co-speech gesture video generation. Our method introduces a weakly-supervised latent deviation module that effectively captures motion features and preserving key appearance details. The results of experiments demonstrate that our approach improve the quality of co-speech gesture video generation in comparison with the state-of-the-art methods.

\bibliographystyle{IEEEbib}
\bibliography{main}

% \vspace{12pt}
% \color{red}
% IEEE conference templates contain guidance text for composing and formatting conference papers. Please ensure that all template text is removed from your conference paper prior to submission to the conference. Failure to remove the template text from your paper may result in your paper not being published.

\end{document}